\documentclass[aps,prl,reprint,amsmath,amssymb,twocolumn]{revtex4-2}
\usepackage{graphicx} 
\usepackage{tikz}
\usetikzlibrary{positioning, arrows.meta}
\usepackage{subcaption}
\usepackage{soul}
\usepackage{listings}
\usepackage{xcolor}
\usepackage{booktabs}
\usepackage{multirow}
\usepackage{array}
\usepackage{xcolor}
\usepackage{float}    
\usepackage{makecell} 
\usepackage{footnote}
\usepackage{CJKutf8}

\usepackage[margin=2.5cm]{geometry} 

\begin{document}

\title{A Status Quo Investigation of Large Language Models towards Cost-Effective CFD Automation with OpenFOAMGPT: ChatGPT vs. Qwen vs. Deepseek}

\author{Wenkang Wang (\begin{CJK*}{UTF8}{gbsn}王文康\end{CJK*})}
\thanks{These authors contributed equally to this work.}
\affiliation{
International Research Institute for Multidisciplinary Science, Beihang University, 100191 Beijing, China
}%

\author{Ran Xu (\begin{CJK*}{UTF8}{gbsn}徐冉\end{CJK*})}
\thanks{These authors contributed equally to this work.}
\affiliation{
Cluster of Excellence SimTech, University of Stuttgart, Stuttgart, Germany
}%

\author{Jingsen Feng (\begin{CJK*}{UTF8}{gbsn}冯晶森\end{CJK*})}
\affiliation{
Faculty of Environment, Science and Economy, University of Exeter, Exeter EX4 4QF, United Kingdom
}%

\author{Qingfu Zhang (\begin{CJK*}{UTF8}{gbsn}张清福\end{CJK*})}
\affiliation{Institute of Fluid Mechanics, 
Beihang University, 100191 Beijing, China
}%

\author{Sandeep Pandey}
\affiliation{
Institute of Thermodynamics and Fluid Mechanics, Technische Universität Ilmenau, Ilmenau D-98684, Germany
}%

\author{Xu Chu (\begin{CJK*}{UTF8}{gbsn}初旭\end{CJK*})}
\email{x.chu@exeter.ac.uk}
\affiliation{
Faculty of Environment, Science and Economy, University of Exeter, Exeter EX4 4QF, United Kingdom
}%
\affiliation{
Cluster of Excellence SimTech, University of Stuttgart, Stuttgart, Germany
}%

\begin{abstract}

We evaluated the performance of OpenFOAMGPT incorporating multiple large-language models. Some of the present models efficiently manage different CFD tasks such as adjusting boundary conditions, turbulence models, and solver configurations, although their token cost and stability vary. Locally deployed smaller models like QwQ-32B struggled with generating valid solver files for complex processes. Zero-shot prompting commonly failed in simulations with intricate settings, even for large models. Challenges with boundary conditions and solver keywords stress the requirement for expert supervision, indicating that further development is needed to fully automate specialized CFD simulations.

\end{abstract}

\maketitle

\newpage

In recent years, the fluid mechanics community has rapidly embraced data-driven strategies, propelled by the proliferation of high-fidelity simulation data and the remarkable progress of machine learning techniques \citep{vinuesa2022enhancing, Duraisamy.2019,cremades2025additive, yang2024data,pandey2020perspective,wang2024optimized}. These methods have shown promise in turbulence modeling, both with and without governing equations \citep{wu2018physics,ling2016reynolds,Beck.2019,chu2024non,yang2019predictive,beck2023toward}, as well as in solving intricate heat transfer problems \citep{Chang.2018,Chu.2018b}. Machine learning has also enhanced various experimental fluid mechanics tasks \citep{vinuesa2023transformative} and facilitated scientific discovery, for example by leveraging causal inference techniques to interpret fluid flow phenomena \citep{Wang.2021,wang2022spatial,liu2023interfacial}.

Simultaneously, Large Language Models (LLMs) such as ChatGPT \citep{achiam2023gpt}, DeepSeek \citep{liu2024deepseek}, Qwen \citep{yang2024qwen2} have surged to the forefront of scientific and engineering research, offering unparalleled capabilities in natural language processing \citep{min2023recent}, automated reasoning \citep{ma2024llm}, and high-level decision-making. Their potential to streamline research pipelines is reflected in diverse applications, from problem-solving and optimization \citep{song2023pre,wang2023prompt,huang2024crispr} to faster scientific breakthroughs \citep{chibwe2024evaluating,ramos2024review}. Notably, these models can serve as intelligent assistants that both amplify traditional methodologies and introduce novel strategies for data analysis, design, and simulation. These advancements are further complemented by work in multimodal vision-language models, exemplified by Cephalo \citep{buehler2024cephalo}, which merges visual and linguistic data to support complex materials design tasks.

Within fluid mechanics, LLMs have proven useful for equation discovery \citep{du2024large} and for streamlining shape optimization \citep{zhang2024usinglargelanguagemodels}, demonstrating efficiency in tasks such as identifying governing equations and refining geometric profiles (e.g., airfoils). Recent research extends LLM-driven insights to microfluidics \citep{xu2024trainingmicrorobotsswimlarge}, where LLMs facilitate decision-making for robotic motion planning, as well as to unsteady flow prediction \citep{zhu2024fluid}, combining pre-trained transformers and graph neural networks for improved spatial-temporal accuracy. In related engineering fields, \citet{kim2024chatgpt} explored ChatGPT for automated MATLAB code generation, proving its utility in providing structured starter codes and logic under expert direction. Likewise, \citet{chen2024metaopenfoam} proposed a multi-agent LLM system for orchestrating CFD workflows via natural language, making these techniques more accessible through retrieval-augmented generation (RAG).

Building on this momentum, our recent work introduced \textbf{OpenFOAMGPT} \citep{pandey2025openfoamgpt}, an LLM-based agent for OpenFOAM-centric CFD simulations that integrates GPT-4o and a chain-of-thought-enabled o1 model. While the o1 model carries a higher token cost—approximately six times that of GPT-4o—it demonstrates consistently superior performance in tasks ranging from zero-shot case setup and boundary condition refinements to turbulence model adjustments and code translation. Through iterative correction loops and RAG for domain-specific knowledge, the framework adeptly handles a range of flow configurations (including single- and multi-phase flow) in just a few iterations, although human oversight remains vital for accuracy and flexibility. Not only does this capability broaden the applicability of CFD methods to non-specialists, but it also signals promising opportunities for adopting LLM-based CFD agents in other solver environments. However, the OpenAI o1 model incurs a relatively high token cost—$\$$15 per million input tokens and $\$$60 per million output tokens—which can become substantial when generating complex setups requiring iterative corrections. Consequently, identifying a more cost-effective solution with comparable performance is highly beneficial.

In this paper, we adapt \textbf{OpenFOAMGPT} to a wider range of LLMs to achieve an more economical agent framework for computational fluid dynamics simulations that leverages two new foundation models—DeepSeek V3 and Qwen 2.5-Max to achieve cost reductions of up to \(\mathcal{O}(10^2)\). Our updated approach seeks to further streamline complex CFD workflows \textit{without the help of RAG}, and scale seamlessly across diverse engineering applications. This evaluates the pure zero-shot performance of OpenFOAMGPT.



OpenFOAM (Open Source Field Operation and Manipulation) is a widely utilized, open-source CFD solver package \citep{weller1998tensorial,Pandey.2018,liu2024simulation}. Unlike commercial alternatives, OpenFOAM's open architecture enables users to modify existing solvers and develop new physical models which is very convenient to combine with LLM. Its C++ object-oriented design ensures maintainability, extensibility, and parallel efficiency, making it valuable for both academic research and industrial applications. These characteristics have positioned OpenFOAM as an effective platform for integration with LLM-based agents that can guide users through simulation workflows. This study employs the OpenFOAM-v2406 release.

\begin{figure}
    \centering
    \includegraphics[width=1\linewidth]{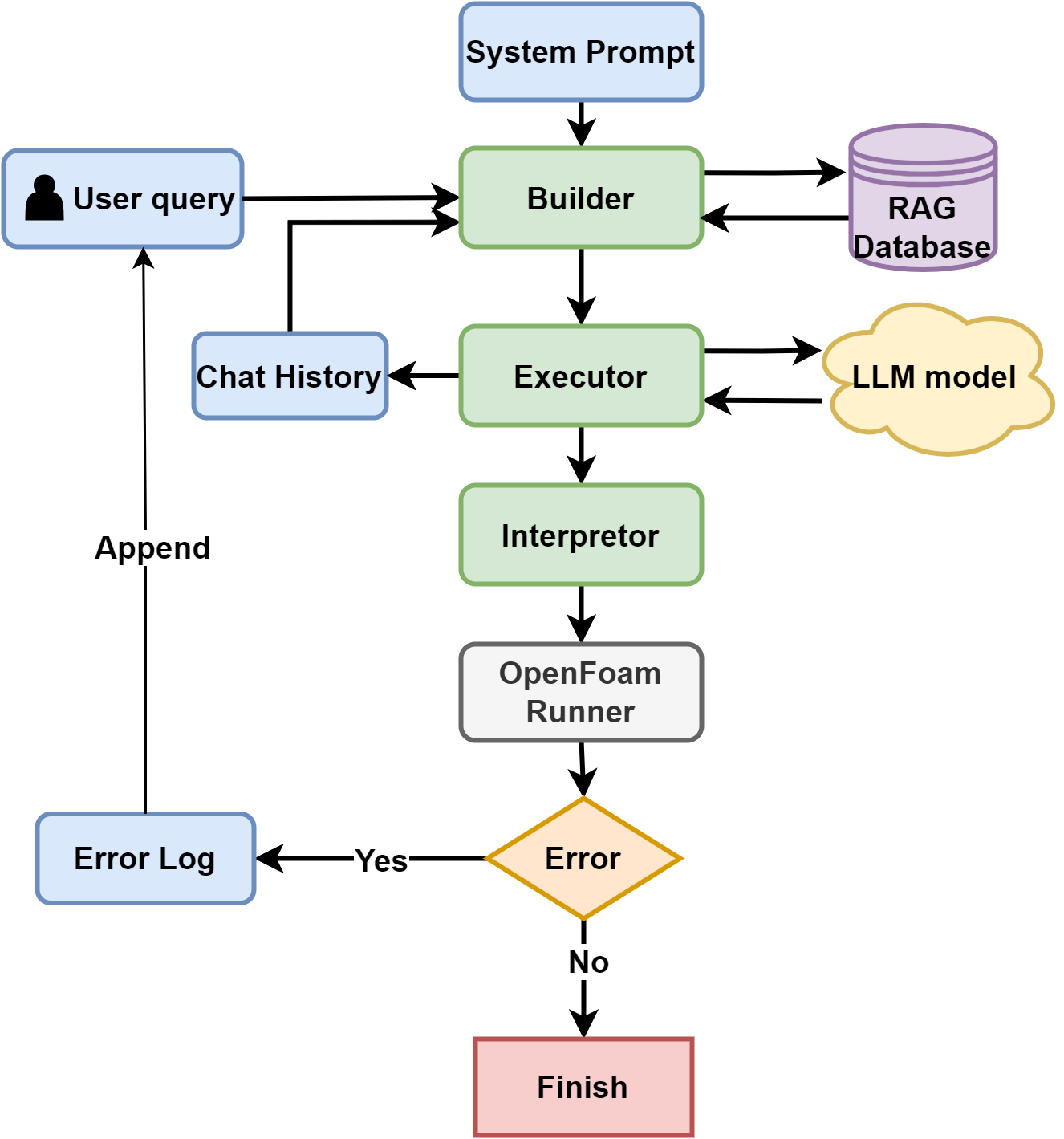}
    \caption{The design of the agent structure}
    \label{fig:agent}
\end{figure}

Figure~\ref{fig:agent} illustrates the hierarchical architecture of our agent \textbf{OpenFOAMGPT}. The workflow begins when a system prompt combines with a user query at the top level. The Builder module then interprets these instructions, consulting the RAG database for domain-specific knowledge when needed, and converts them into a structured execution plan. Subsequently, the Executor manages the workflow by either directing queries to the LLM model for additional reasoning or delegating tasks to the Interpreter that translate the output to files needed for simulation operations. OpenFOAM runner then executes the simulation with the setup files, and the system output and error logs are continuously monitored during simulation; upon failure detection, the error data are appended to the original query and the process cycles again. Otherwise, the workflow terminates successfully. Unlike our previous study, the RAG function is disabled for the presented research to test the pure zero-shot capability.

The following 4 LLMs are considered and evaluated currently. 

\begin{itemize}
    \item  \textbf{ChatGPT-4o} is a general-purpose multi-modal LLM developed by OpenAI. Trained on a diverse range of internet text, it maintains advanced language understanding and generation capabilities across domains.
    
    \item \textbf{OpenAI o1} is the first reasoning model leveraging a chain-of-thought (CoT) mechanism, enabling superior performance over GPT-4o in complex reasoning, scientific analyses, and programming tasks.
    
    \item \textbf{DeepSeek V3 (671B)} is the third-generation LLM from DeepSeek AI, offered as an open-source alternative to high-end proprietary models. Although DeepSeek R1 incorporated a reasoning engine based on DeepSeek V3, our evaluations indicated that DeepSeek V3 outperforms DeepSeek R1 in most scenarios. 
    
    \item \textbf{Qwen2.5-Max} is Alibaba's latest LLM, designed as a general-purpose MoE AI system with great performance on various standard tests for LLM. 
\end{itemize}

The comparison of token pricing across LLM substantial economic differentials between US and Chinese providers (Table \ref{tab:llm_comparison}). The OpenAI o1 model represents the highest-cost option at \$15.0 and \$60.0 per million tokens for input and output processing, respectively, while offering the most extensive context window (200k tokens). In contrast, DeepSeek-V3 671B demonstrates remarkable cost efficiency at \$0.035 and \$0.55 per million tokens—approximately $10^2$ less expensive than o1 for equivalent token processing. GPT-4o presents an intermediate pricing tier among US-based models (\$2.5/\$10.0 per million input/output tokens) with a 128k context window. The Qwen 2.5-Max model maintains competitive pricing at \$0.80 and \$1.2 per million tokens despite its more limited 32k context capacity. These pricing differentials substantiate our framework's approach of leveraging Chinese models to achieve the $\mathcal{O}(10^2)$ cost reduction claimed in our computational fluid dynamics agent implementation while maintaining acceptable performance characteristics.

\begin{table*}[htbp]
\footnotesize
\centering
\caption{Comparison of used large language models}
\label{tab:llm_comparison}
\begin{tabular}{@{}lcccccc@{}}
\toprule
\textbf{Model} & \makecell{\textbf{Input price} \\ (\$/1M tokens)} & \makecell{\textbf{Output price} \\ (\$/1M tokens)}& \textbf{\makecell{Context \\length}}  & \textbf{Country}  \\ 
\midrule
GPT-4o & 2.5 & 10.0 & 128k  & USA  \\
OpenAI o1 & 15.0 & 60.0 & 200k  & USA  \\
DeepSeek-V3$^{***}$ & 0.035 & 0.55 & 64k  & China \\
Qwen 2.5-Max & 0.80 & 1.2 & 32k  & China  \\
\bottomrule
\end{tabular}

\footnotesize
*Prices are approximate and based on public API rates (2025 Q1).\\
**Context window sizes may vary by implementation.\\
***Discount price (UTC 16:30-00:30)
\end{table*}


The local deployment of large language models (LLMs) currently relies on three mainstream frameworks: SGLang \citep{sglang}, vLLM \citep{kwon2023pagedattention}, and Ollama \citep{ollama}  . While SGLang and vLLM emphasize multi-GPU parallelization for industrial-scale inference, Ollama's lightweight architecture (v0.4.7) demonstrates superior suitability for single-GPU consumer hardware. We deployed the QwQ-32B Q4\_K\_M quantized model (4-bit precision with medium granularity) on an NVIDIA RTX 4090 GPU (24GB VRAM), achieving stable operation with 20.3GB VRAM utilization. Integration into the OpenFOAMGPT workflow was implemented through local API calls (port 11434), enabling direct interaction with CFD simulation templates.

Testing encompassed two benchmark cases: the lid-driven Cavity flow (laminar regime) and PitzDaily combustor flow (turbulent reactive case). Both scenarios were evaluated under zero-shot prompting and retrieval-augmented generation (RAG) conditions, using OpenFOAM v2406 documentation as supplementary context. Despite 20 maximum inference iterations per trial, neither configuration successfully generated valid solver files adhering to OpenFOAM syntax requirements. Notably, identical failure patterns emerged when querying the official non-quantized QwQ-32B via API, eliminating quantization artifacts as the primary failure cause.

These results align with recent theoretical analyses of domain adaptation thresholds \citep{chowdhery2023palm} , where sub-100B parameter general-purpose models exhibit critical knowledge gaps in specialized engineering domains. Error analysis revealed systematic failures in boundary condition specification and turbulence model parameterization \citep{brunton2020fluid}. Such limitations suggest that effective deployment in computational mechanics workflows requires either: scaling to foundation models, or domain-specific fine-tuning of smaller architectures.



Zero-shot prompting attempts to generate desired output based upto limited instructions without any example. As LLMs are trained on vast datasets, therefore, it often results in desired outputs. However, if underlying problem is difficult then model might not provide desired output. In this subsection, we analyze the performance of LLM models with zero-shot prompting. The evaluation encompasses a range of typical CFD engineering tasks, including modifying initial and boundary conditions, adjusting turbulence models, and updating thermophysical properties, as listed in TABLE \ref{BCIC}.


\begin{itemize}
    \item \textbf{Cavity flow:} Simulates laminar, isothermal, incompressible flow in a square cavity using \texttt{icoFoam}. The top wall moves horizontally at 1~m/s; other walls are stationary.
    
    \item \textbf{PitzDaily:} Models incompressible turbulent flow through a two-dimensional sudden expansion channel using the k-$\epsilon$ turbulence model and \texttt{simpleFoam} solver.
    
    \item \textbf{Hotroom:} Simulates turbulent natural convection in a tall rectangular cavity using the k-$\epsilon$ model and \texttt{buoyantBoussinesqSimpleFoam}. The bottom wall is heated, the top wall cooled, and side walls are adiabatic.
    
    \item \textbf{Dambreak:} Represents a simplified laminar dam break using the VOF-based \texttt{interFoam} solver. A water column collapses into a square tank containing a central rectangular obstacle, creating complex flow patterns and trapped air pockets.
    
    \item \textbf{Particle column:} Uses \texttt{MPPICFoam} to simulate particle dynamics and fluid flow in a vertical rectangular column. Fluid motion is described by Navier-Stokes equations, and particles are tracked via the Lagrangian approach, considering drag, collisions, and gravity.
    
    \item \textbf{Mixed vessel:} Simulates fluid mixing in a rotary agitator using \texttt{pimpleFoam}. The geometry features a cylindrical domain with rotating inner walls, stationary outer walls, and rectangular barriers to enhance mixing.
\end{itemize}

\begin{table}[htbp]
    \centering
    \footnotesize
    \begin{tabular}{|>{\centering\arraybackslash}m{0.18\linewidth}|
                    >{\centering\arraybackslash}m{0.22\linewidth}|
                    >{\centering\arraybackslash}m{0.20\linewidth}|
                    >{\centering\arraybackslash}m{0.065\linewidth}|
                    >{\centering\arraybackslash}m{0.060\linewidth}|
                    >{\centering\arraybackslash}m{0.085\linewidth}|
                    >{\centering\arraybackslash}m{0.073\linewidth}|}
    \hline
    Case  & \multicolumn{2}{c|}{Alternate condition} & 4o & o1 & Qwen & DS\\
    \hline
    \multirow{8}{*}{Cavity flow} 
     & Top wall velocity & 1m/s $\to$ 2m/s & $\checkmark$ &  & $\checkmark$ & $\checkmark$ \\
    \cline{2-7}
     & Top wall velocity & 1m/s $\to$ $5\sin\bigl(2\pi \frac{t}{0.1}\bigr)$ & $\times$ & $\checkmark$ & $\times$ & $\times$ \\
    \cline{2-7}
     & Mesh resolution & 20$\times$20$\times$1 $\to$ $\times$15$\times$1 & $\checkmark$ &  & $\checkmark$ & \\
    \cline{2-7}
     & endTime & 3 $\to$ 5 & $\checkmark$ &  & $\checkmark$ & $\checkmark$ \\
    \cline{2-7}
     & Turbulence model & RNGkepsilon & $\checkmark$ &  & $\times$ & \\
    \cline{2-7}
     & Turbulence model & kOmegaSST & $\checkmark$ &  & $\checkmark$ & \\
    \cline{2-7}
     & Turbulence model & kkLOmega & $\times$ & $\checkmark$ & $\times$ & \\
    \cline{2-7}
     & Turbulence model & LRR & $\checkmark$ &  & $\times$ & \\
    \hline
    \multirow{3}{*}{PitzDaily}
     & Inlet velocity & 10m/s $\to$ 20m/s & $\checkmark$ &  & $\checkmark$ & $\times$ \\
    \cline{2-7}
     & Turbulence model  & kOmegaSST & $\checkmark$ &  & $\times$ & \\
    \cline{2-7}
     & Turbulence model  & Smagorinsky (LES) & $\checkmark$ &  & $\checkmark$ & \\
    \hline
    Hotroom 
     & HOT\_WALL temperature & 310K $\to$ 320K & $\checkmark$ &  & $\checkmark$ & $\checkmark$ \\
    \hline
    \multirow{2}{*}{Dambreak}
     & Liquid inside the membrane & water $\to$ oil & $\checkmark$ &  & $\checkmark$ & $\times$ \\
    \cline{2-7}
     & Turbulence model  & KEpsilon & $\checkmark$ &  & $\times$ & \\
    \hline
    \multirow{3}{*}{\makecell{Particle \\ column} }
     & Velocity of the fluid/particles & 1m/s $\to$ 2m/s & $\checkmark$ &  & $\checkmark$ & $\checkmark$ \\
    \cline{2-7}
     & Type of fluid & Air $\to$ CO & $\times$ & $\checkmark$ & $\checkmark$ & $\checkmark$ \\
    \cline{2-7}
     & Turbulence model & KEpsilon & $\times$ & $\times$ & $\times$ & \\
    \hline
    \multirow{2}{*}{\makecell{Mixed \\ vessel}}
     & Angular speed of rotation & 20rad/s $\to$ 15rad/s & $\checkmark$ &  & $\times$ & $\times$ \\
    \cline{2-7}
     & \makecell{Turbulence \\Model }& KEpsilon & $\times$ & $\checkmark$ & $\times$ & \\
    \hline
    \end{tabular}
    \caption{Tasks of alternate initial- and boundary conditions. o1 were only tested when gpt-4o failed}
    \label{BCIC}
\end{table}

From TABLE \ref{BCIC}, Qwen2.5-Max delivers performance on par with OpenAI o1, while dramatically reducing token costs. Notably, the reasoning model DeeSeek R1 performed worse than DeepSeek V3—quite the opposite of the improvement observed when switching from OpenAI gpt-4o to o1. Additionally, it is worth mentioning that the direct API output from DeepSeek V3 tends to be unstable, and the output from the third-party API falls short of the quality produced by the original API of Deepseek.


We further evaluated Qwen2.5-Max’s ability to generate and debug simulations, encompassing a series of classical single- and multi-phase scenarios.

\begin{figure}
    \centering
    \includegraphics[width=1.1\linewidth]{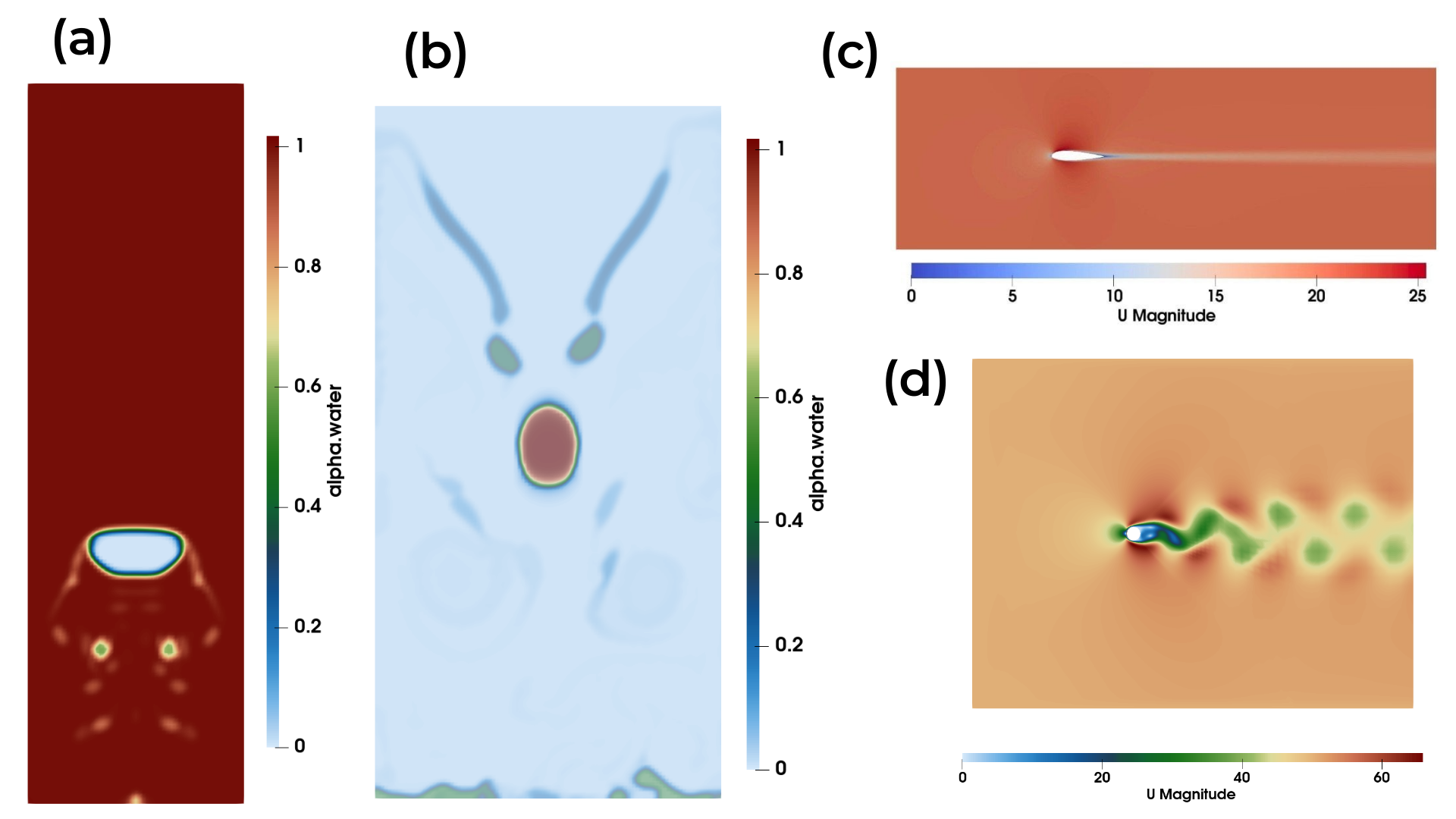}
    \caption{Case simulation results. (a) 2D rising bubble (b) 2D falling droplet (c) airFoil2D (d) cylinder.}
    \label{result2}
\end{figure}

\begin{itemize}

    \item \textbf{2D rising bubble} (FIG.\ref{result2}(a)): The setup consists of a rectangular tank filled with water, measuring 30 mm in width and 100 mm in height. Initially, a bubble with a diameter of 10 mm is positioned centrally at the bottom of the domain. Buoyancy-driven motion induces the bubble to rise, deform, and interact dynamically with the surrounding fluid. 

    \item \textbf{2D falling droplet} (FIG.\ref{result2}(b)): This case examines the dynamics of a single water droplet falling under gravity within a two-dimensional rectangular tank filled with air, employing the Volume of Fluid (VoF) method. Initially, at \( t = 0 \) s, the droplet is positioned centrally at the tank's upper boundary and subsequently descends.

    \item \textbf{2D airfoil} (FIG.\ref{result2}(c)): This case investigates the aerodynamic performance of a two-dimensional NACA 0012 airfoil positioned at a 5° angle of attack within a computational wind tunnel. The domain dimensions are 2000 mm in length and 1000 mm in width. At the simulation start (\( t = 0 \) s), a uniform airflow at 20 m/s enters the domain, initiating steady-state conditions driven by pressure gradients and viscous forces interacting with the airfoil surface. 

    \item \textbf{3D MotorBike}: This case investigates aerodynamic and turbulent flow characteristics around a simplified motorBike geometry. The transient airflow interaction is modeled for a motorcycle body with overall dimensions of 2.1 m × 0.8 m × 1.2 m. The three-dimensional computational domain extends 20 vehicle lengths upstream and downstream, placing the motorcycle 10 m downstream from the inlet boundary.
    
    \item \textbf{2D cylinder} (FIG.\ref{result2}(d)): This case examines aerodynamic and vortex-induced phenomena around a two-dimensional circular cylinder. The cylinder is centered at the origin within a rectangular computational domain featuring clearly defined boundaries: inlet (left side), outlet (right side), and walls (top and bottom). Initially, at \( t = 0 \) s, a uniform freestream velocity of 1 m/s is imposed.

    \item \textbf{2D nozzleFlow2D}: This case investigates axisymmetric high-speed fuel injection. The computational domain includes a 3 mm diameter inlet connected to a gradually expanding throat. At the initial time \( t = 0 \) s, diesel fuel is injected at a velocity of 460 m/s into a low-pressure gas environment maintained at atmospheric conditions. The simulation employs the Volume of Fluid (VoF) method coupled with a Large Eddy Simulation (LES) turbulence model. 
    
\end{itemize}

\begin{table}[htbp]
  \centering
  \caption{Evaluations with Qwen2.5-Max}
    \begin{tabular}{|c|c|c|>{\centering\arraybackslash}m{0.25\linewidth}|
                    >{\centering\arraybackslash}m{0.1\linewidth}|c|}
    \hline
    case  & file provided & iterations & result & total token & \makecell{token \\cost} \\
    \hline
    Bubble & \makecell{ \texttt{blockMeshDict}, \\ \texttt{setFieldsDict}} & 8     & $\surd$ & 71k & \$0.25 \\
    \hline
    Droplet & \makecell{ \texttt{blockMeshDict}, \\ \texttt{setFieldsDict} } & 20    & $\surd$ & 195k & \$0.67\\
    \hline
    AirFoil & \texttt{polyMesh} & 2     & $\surd$ & 15k & \$0.056\\
    \hline
    MotorBike & \texttt{polyMesh} &   10    & patch type 'patch' not constraint type 'empty' & 66k & \$0.23 \\
    \hline
    Cylinder & \texttt{polyMesh}   &   3    &    $\surd$   & 15k & \$0.05 \\
    \hline
    Nozzle & \texttt{blockMeshDict}  & 20    & 'smoother' not found in "fvSolution" & 127k & \$0.37 \\
    \hline
    \end{tabular}%
  \label{tab:addlabel}%
\end{table}%

Table \ref{tab:addlabel} shows the results of the computational experiments. The results confirm that Qwen 2.5-Max can handle selected CFD cases without needing RAG support. When given geometric models and mesh files, the LLM can generate and debug simulations for rising bubble, falling droplet, airfoil, and cylinder cases.
More complex cases, such as motorBike and nozzleFlow, show additional challenges. For the motorBike simulation, Qwen 2.5-Max exhibited persistent failures in boundary condition configuration despite explicit guidance, with repeated attempts to implement 'patch' or 'empty' boundary types proving unsuccessful. For nozzleFlow, error messages showed a need for 'smoother' entries, a minor issue for experienced engineers. However, after being specifically told to make these corrections, the system still had setup errors. It kept missing critical \texttt{p\_rgh} file dependencies that were clearly listed in the workflow requirements.

We extended OpenFOAMGPT by integrating two significantly more affordable large language models, DeepSeek V3 and Qwen 2.5-Max, achieving cost savings of up to two orders of magnitude compared to OpenAI o1. These models are proved capable of handling common CFD tasks—such as modifying boundary conditions, turbulence models, and solver configurations—across a variety of test cases. We also explored a locally deployed QwQ-32B model running on a single desktop GPU, though it struggled to produce fully correct solver files for specialized engineering tasks, suggesting that smaller models may need domain-specific training or fine-tuning to handle complex scenarios.

Despite these encouraging results, certain challenges persist. In more intricate simulations—for instance, cases with complex geometry setups—zero-shot prompting alone often fell short. Repeated boundary-condition errors and missing solver keywords highlight the need for human oversight or additional AI guidance, especially when dealing with less-documented features of OpenFOAM. Though the less expensive models successfully reduced token costs, they sometimes suffered from narrower context windows and struggled with multi-step error correction, indicating room for improvement in handling elaborate CFD workflows.

Looking ahead, several avenues seem promising. First, smaller or mid-sized models could be fine-tuned using specialized CFD corpora to boost accuracy while keeping inference costs low. Second, bridging textual instructions with geometry and mesh data remains a hurdle—multi-modal approaches and more sophisticated prompt-engineering strategies could help LLMs interpret problem setups in a more intuitive way. Third, carefully combining zero-shot techniques with retrieval-augmented generation may offer a practical blend of lower costs and more reliable outcomes. Lastly, improving local deployment on consumer-grade GPUs—whether by scaling up model sizes or refining quantization—could reduce dependence on external APIs. By pursuing these directions, we can inch closer to creating robust, flexible, and truly cost-effective LLM-driven CFD solutions for both research and industry.

\section*{Acknowledgments}

XC appreciates the funding support from Royal Society (RG\textbackslash R1\textbackslash 251236). WW is supported by the Fundamental Research Funds for the 
Central Universities (China).

\bibliography{sn-bibliography}

\end{document}